# APTx: better activation function than MISH, SWISH, and ReLU's variants used in deep learning

Ravin Kumar [0000-0002-3416-2679]

Department of Computer Science, Meerut Institute of Engineering and Technology, meerut-250005, Uttar Pradesh, India
ravin.kumar.cs.2013@miet.ac.in

**Abstract.** Activation Functions introduce non-linearity in the deep neural networks. This nonlinearity helps the neural networks learn faster and efficiently from the dataset. In deep learning, many activation functions are developed and used based on the type of problem statement. ReLU's variants, SWISH, and MISH are goto activation functions. MISH function is considered having similar or even better performance than SWISH, and much better than ReLU. In this paper, we propose an activation function named APTx which behaves similar to MISH, but requires lesser mathematical operations to compute. The lesser computational requirements of APTx does speed up the model training, and thus also reduces the hardware requirement for the deep learning model. Source code: https://github.com/mr-ravin/aptx_activation

## 1 Introduction

The ability of deep learning models to learn features directly from the data have made it a default approach to solve many complex problems. A simple artificial neuron is linear in nature, also expressed in Equation 1.

$$y = \sum w_i x_i + b \tag{1}$$

Here,

y is output from the neuron
$x_i$ is the input to the neuron
$w_i$ is the associated weights
b is the associated bias

When the output of this neuron is passed to an activation function the nonlinearity gets introduced in the network. When considering an activation function one important thing is that the derivative of an activation function should not be the same in its domain. Generally, activation function **f** is applied to the output of the neurons in the hidden layers to make the neural network learn complex features as expressed in Equation 2.

$$\text{output} = \mathbf{f}(y) \tag{2}$$

The SWISH activation function is considered better than the ReLU function and its variants. But, recently developed activation function MISH is considered equivalent or even better than SWISH activation function in some cases.

In this paper, we propose an activation function APTx which behaves similar to the MISH activation function but requires lesser mathematical operations. It means lesser computation is required in APTx to calculate output in the forward propagation, as a result significantly reducing the hardware requirements for training and inference phases. The derivative of APTx also has lesser operations than MISH, hence making neural networks train faster compared to MISH activation function.

## 2 Related Works

Vinod Nair et al. [1] studied the effect of rectified linear units (ReLU) on Restricted Boltzmann Machines. Abien Fred M. Agarap [2] made use of ReLU with convolutional neural networks on the MNIST dataset which outperformed the CNN with softmax on classification task. Glorot et al. [3] and Sun et al. [4] discussed the sparsity of ReLU as a reason for its better performance. Szandała, Tomasz et al. [5] performed a comparative analysis showing tanh and sigmoid function both having vanishing gradient problems overcome by ReLU, and showing the dying-ReLU problem for negative values. Mass et al. [6] presented an improved version of ReLU called Leaky-ReLU where instead of having zero value for negative input the function will have some negative number output. Clevert et al. [7] proposed an ELU function that was faster and better than both ReLU and Leaky-ReLUs. Ramachandran P et al [8] presented SWISH activation function having superior performance than ReLU and its variants. Misra D. et al. [9] proposed an activation function MISH having similar, and in some cases even better performance than SWISH activation function.

## 3 Proposed APTx activation function

We are proposing an activation function named as "Alpha Plus Tanh Times" or APTx in short. Our APTx function is presented as $\boldsymbol{\phi}$ in Equation 3, and its derivative is shown in Equation 4.

$$\boldsymbol{\phi}(x) = (\boldsymbol{\alpha} + tanh(\beta x)) * \gamma x \quad (3)$$

$$\boldsymbol{\phi}'(x) = \gamma\,(\boldsymbol{\alpha} + tanh(\beta x) + \beta x\, sech^2(\beta\, x)) \quad (4)$$

By updating the values of the parameters $\boldsymbol{\alpha}$, $\beta$, and γ we can make the function $\boldsymbol{\phi}$ behave like a MISH activation function (i.e., $\boldsymbol{\alpha}$, $\beta$, and γ can be configured as trainable parameters or specified with static, fixed values). The updated function $\boldsymbol{\phi}$ and its derivative is shown in Equation 5 and 6, where $\boldsymbol{\alpha} = 1$, $\beta = 1$ and $\gamma = ½$ .

$$\boldsymbol{\phi}(x) = (1 + tanh(x)) * x/2 \quad (5)$$

$$\boldsymbol{\phi}'(x) = (1 + tanh(x) + x\, sech^2(x))/2 \quad (6)$$

For the detailed visual analysis of the behavior of our APTx its graph is shown in Figure 1, and the graph of its derivative is shown in Figure 2.

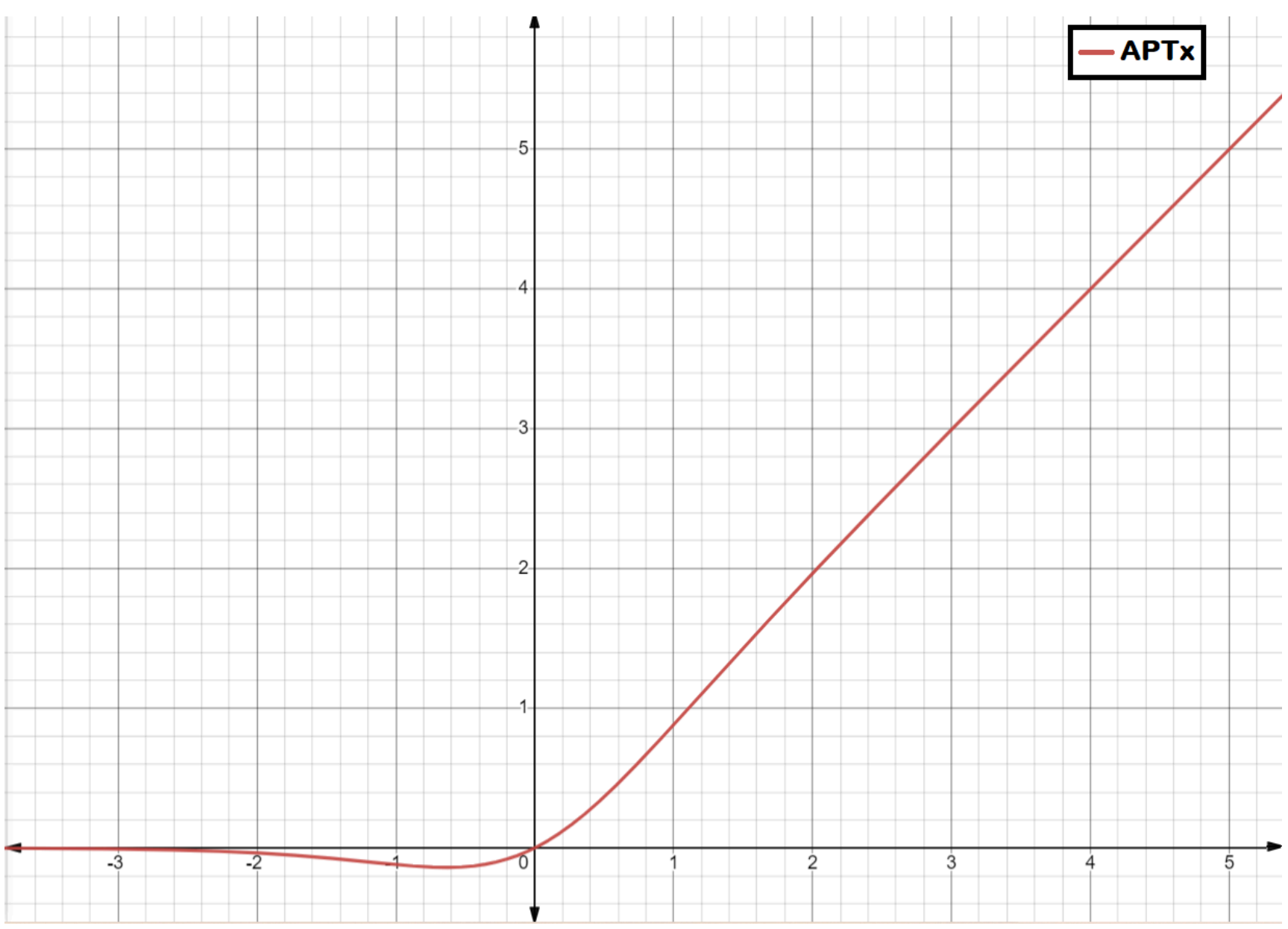

**Figure 1**: Graph of our proposed APTx activation function at $\alpha = 1$, $\beta = 1$ and $\gamma = ½$

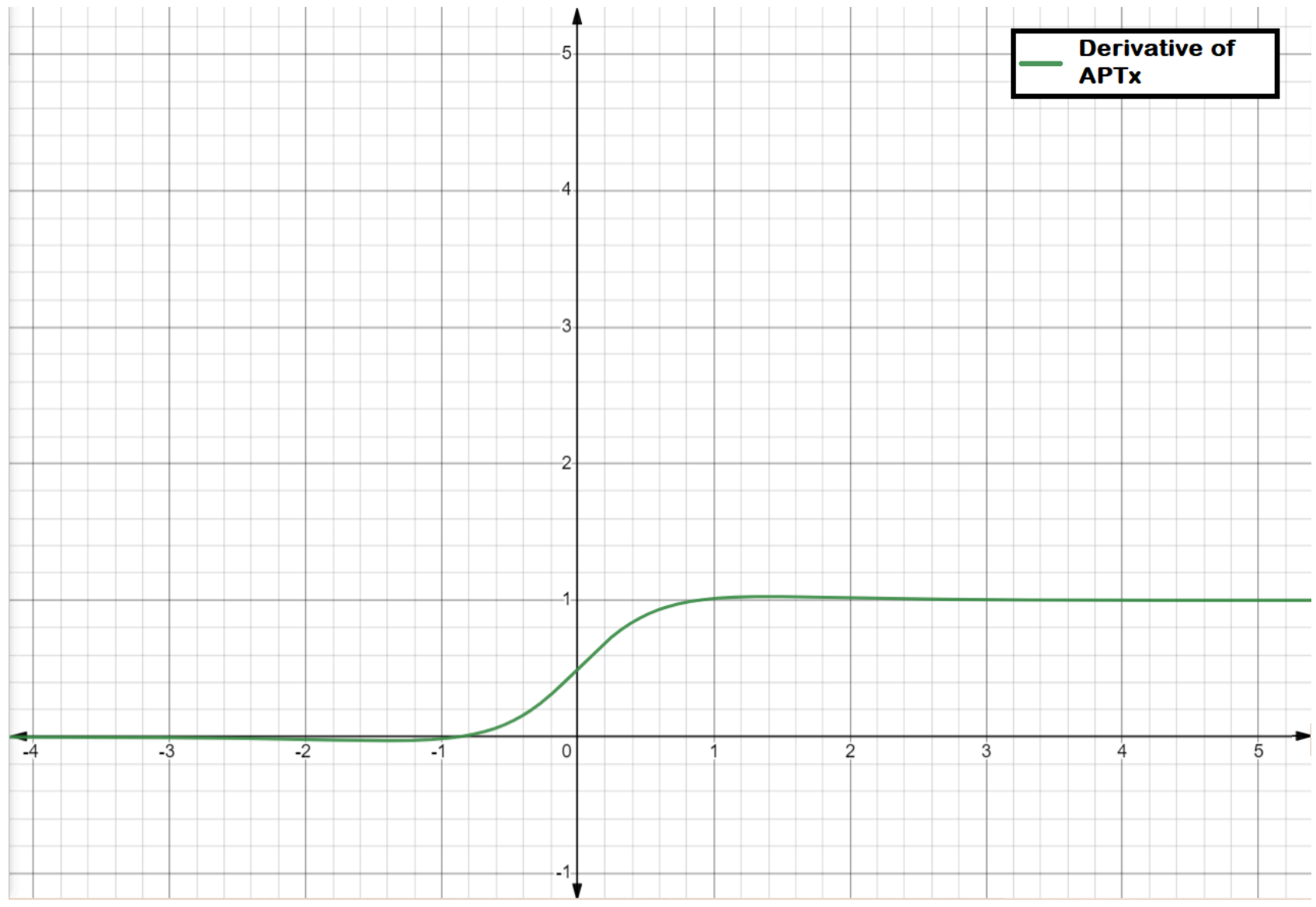

**Figure 2**: Graph of the derivative of APTx activation function at $\alpha = 1$, $\beta = 1$ and $\gamma = ½$

Although one decides activation functions based on the type of the problem statement, there are some popular activation functions whose comparisons were already done in existing research works. First, we discuss how MISH activation function is better than SWISH, ELU, Leaky-ReLU, ReLU, Tanh and Sigmoid activation function for general scenarios. Afterwards, we compared the MISH activation function with our proposed APTx function.

## 4 Comparative analysis of existing activation functions

The sigmoid activation function is mathematically expressed in Equation 7, and comparison of its derivative with the derivative of tanh is shown in Figure 3. One can easily notice in Figure 3 that the range of tanh derivatives is larger than sigmoid derivatives, but for numbers away from zero both tanh and sigmoid have very less output, this introduces the Vanishing Gradient Problem [5] in the larger neural networks.

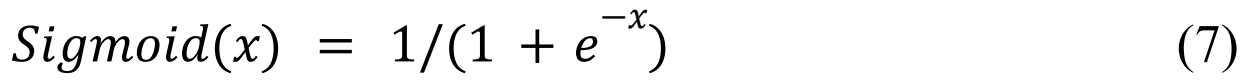

$$Sigmoid(x) = 1/(1 + e^{-x}) \quad (7)$$

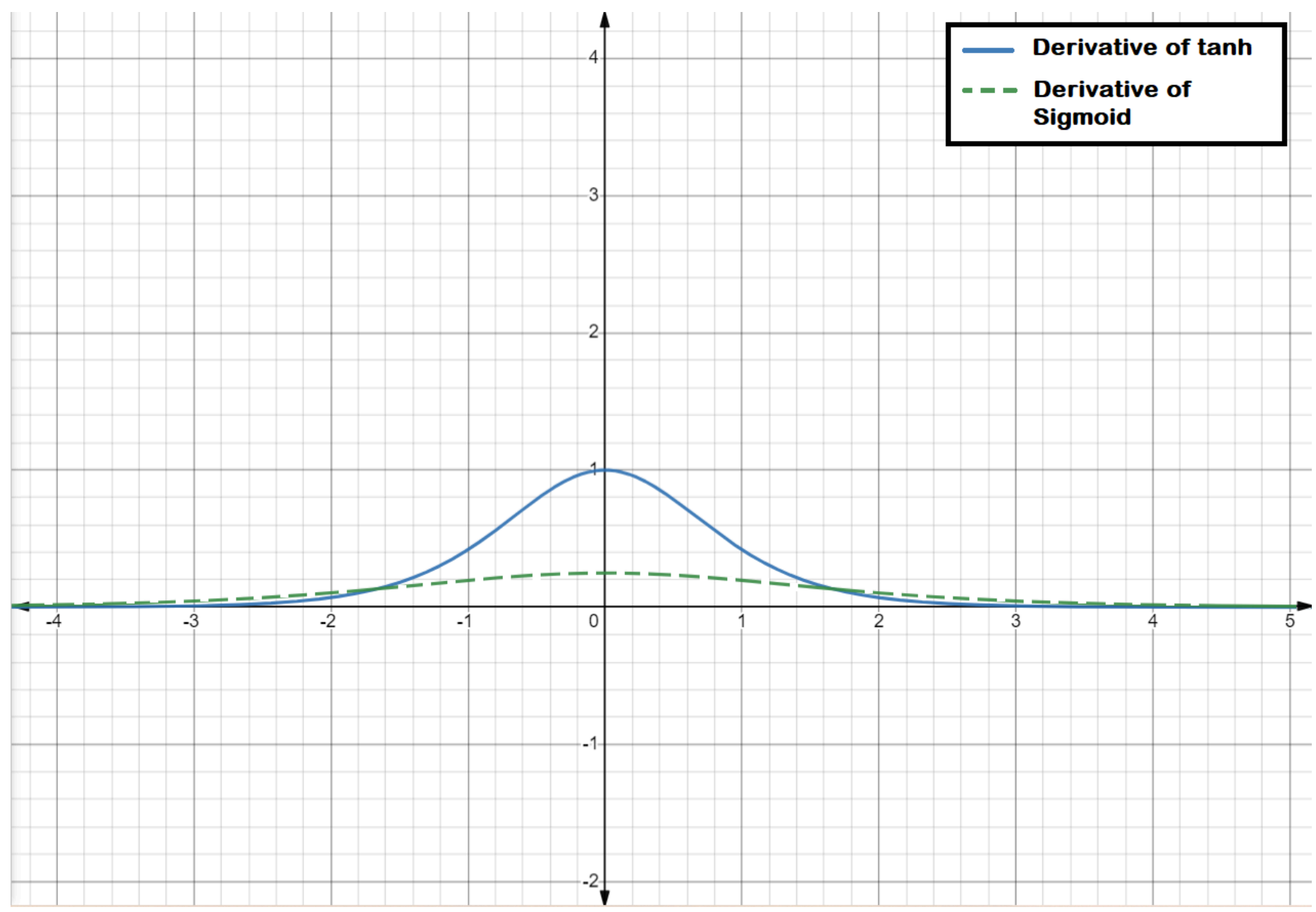

**Figure 3**: Graph showing derivatives of tanh and sigmoid activation functions

The ReLU activation function provided a solution to the Vanishing Gradient Problem at least for the positive inputs [3-4], but for the negative inputs it suffers from the Dying-ReLU problem [5], as its derivative for negative value is Zero. Leaky-ReLU [6] was able to solve the Dying-ReLU problem upto some extent. ELU [7] showed better performance than Leaky-ReLU in most of the tasks as it tends to converge cost to zero faster and produce accurate results. For the positive input ReLU,

Leaky-ReLU, and ELU all behave in the same manner, but the difference lies for the non-positive values as shown in Figure 4 and also presented in Equations 8, 9, and 10 respectively.

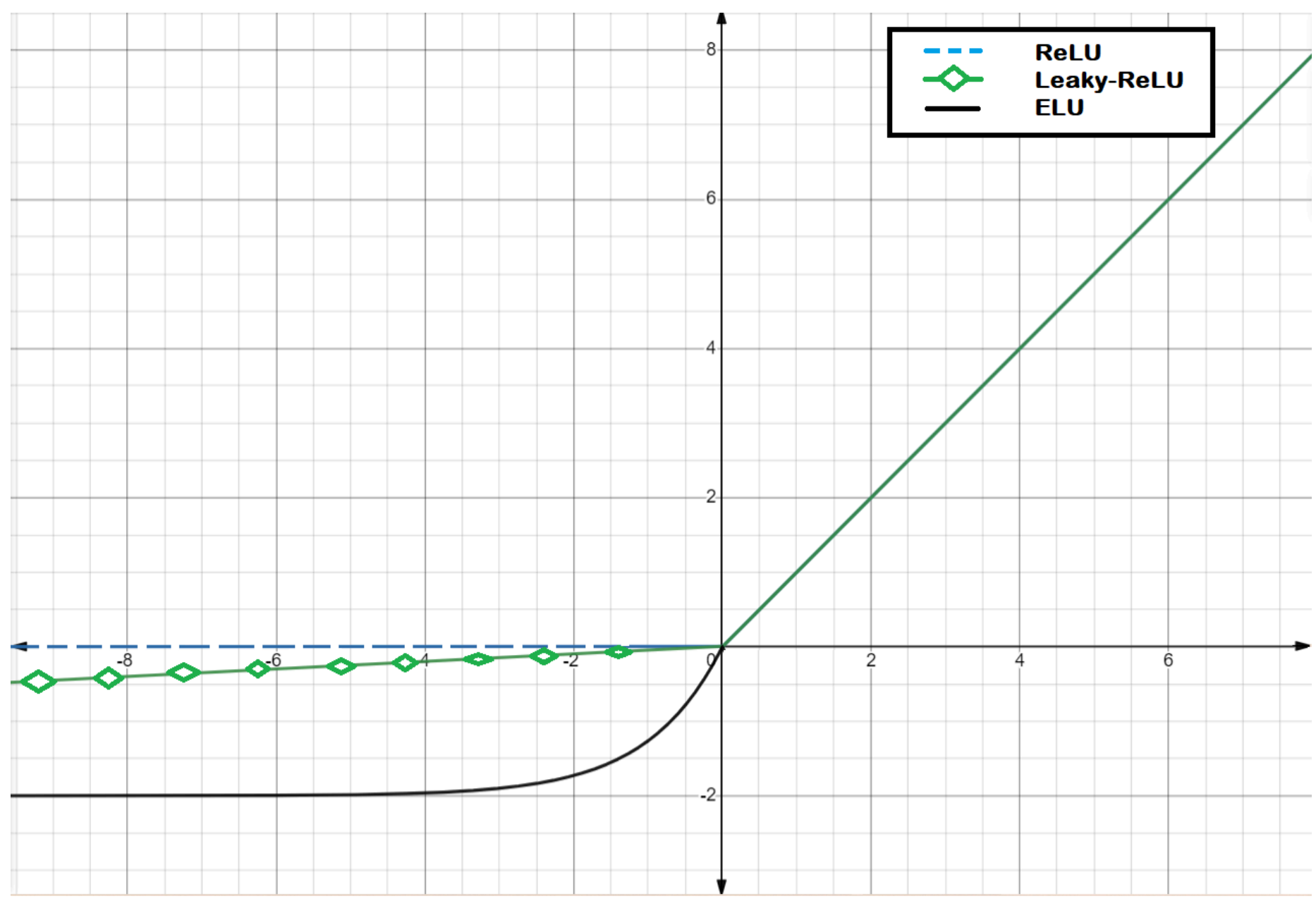

**Figure 4**: Graph of ReLU, Leaky-ReLU (with $\alpha = 0.05$), and ELU (with $\alpha = 2$)

$$ReLU(x) = max(0, x) \tag{8}$$

$$LeakyReLU(x) = \{ \alpha x, \quad x <= 0 \}, \text{ and } \{x, \quad x > 0\} \tag{9}$$

$$ELU(x) = \{ \alpha(e^{x} - 1), \quad x <= 0 \}, \text{ and } \{x, \quad x > 0\} \tag{10}$$

SWISH activation function [8] performs better than ReLU activation function, and also its variants because none of these variants have managed to replace the inconsistent gains (i.e. calculation of derivatives). SWISH can be considered a type of self-gated function, also expressed in Equation 11.

$$SWISH(x) = x * Sigmoid(x) \tag{11}$$

Although introduction of SWISH solved both vanishing gradient and providing consistent gains, development of MISH activation function [9] turned out to provide equivalent and in many tasks it had even better performance than SWISH activation function. Its mathematical form is presented in Equation 12.

$$MISH(x) = x * tanh(ln(1 + e^{x})) \tag{12}$$

Graphs of the derivatives of SWISH and MISH functions are plotted in Figure 5.

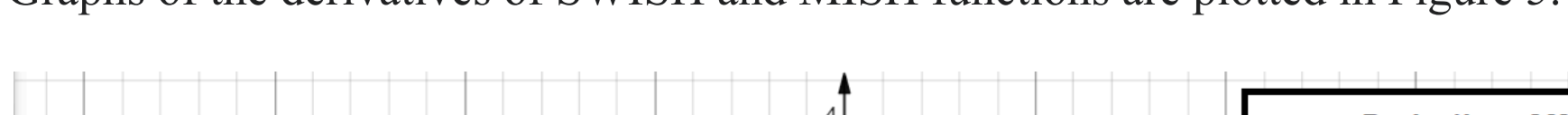

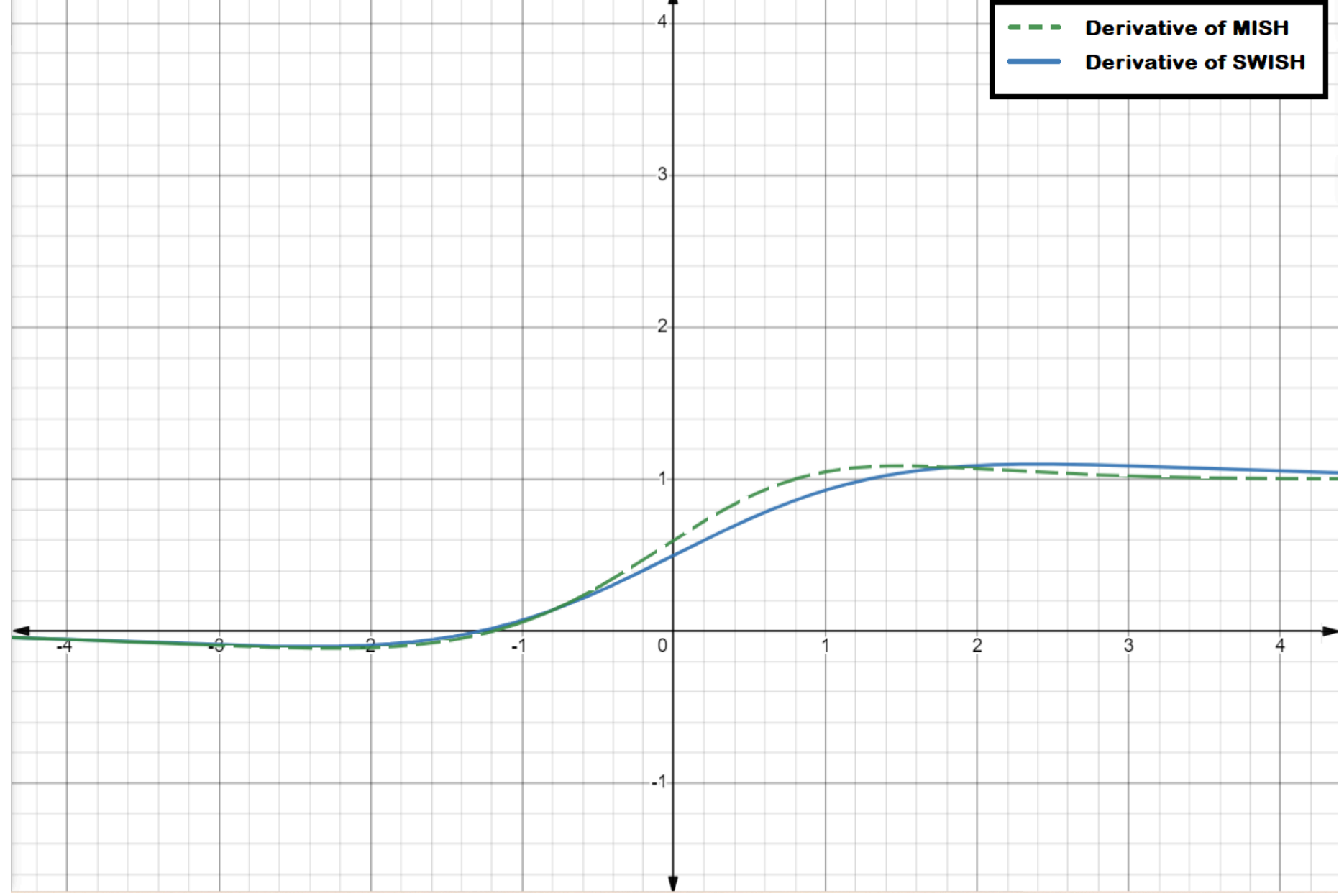

**Figure 5**: Graph of the derivatives of SWISH and MISH activation functions

## 5 Comparative analysis of MISH with proposed APTx

During the forward propagation the mathematical operations required to calculate APTx expressed in Equation 5 are lesser than the MISH activation function shown in Equation 12. But, similar to the MISH function, APTx is bounded below and unbounded above.

The biggest advantage of APTx lies during the training phase while performing backpropagation. Back Propagation requires calculation of derivatives for each epoch and APTx requires fewer mathematical operations to compute its derivative than the MISH activation function. The derivative of MISH is expressed in Equation 13, and for comparative analysis the derivative of APTx is stated again in the Equation 14, where $\boldsymbol{\alpha} = 1$, $\beta = 1$ and $\gamma = ½$.

$$MISH'(x) = (e^{x}(4(x + 1) + 4e^{2x} + e^{3x} + e^{x}(4x + 6)))/(2e^{x} + e^{2x} + 2)^{2} \quad (13)$$

$$\boldsymbol{\phi}'(x) = (1 + tanh(x) + x\,sech^{2}(x))/2 \quad (14)$$

Interestingly, despite the fact that the derivative of the APTx function requires fewer operations than the derivative of MISH and also SWISH. The derivative graphs of APTx and MISH are presented in Figure 6 showing similar behavior for the positive domain part, useful for backpropagation.

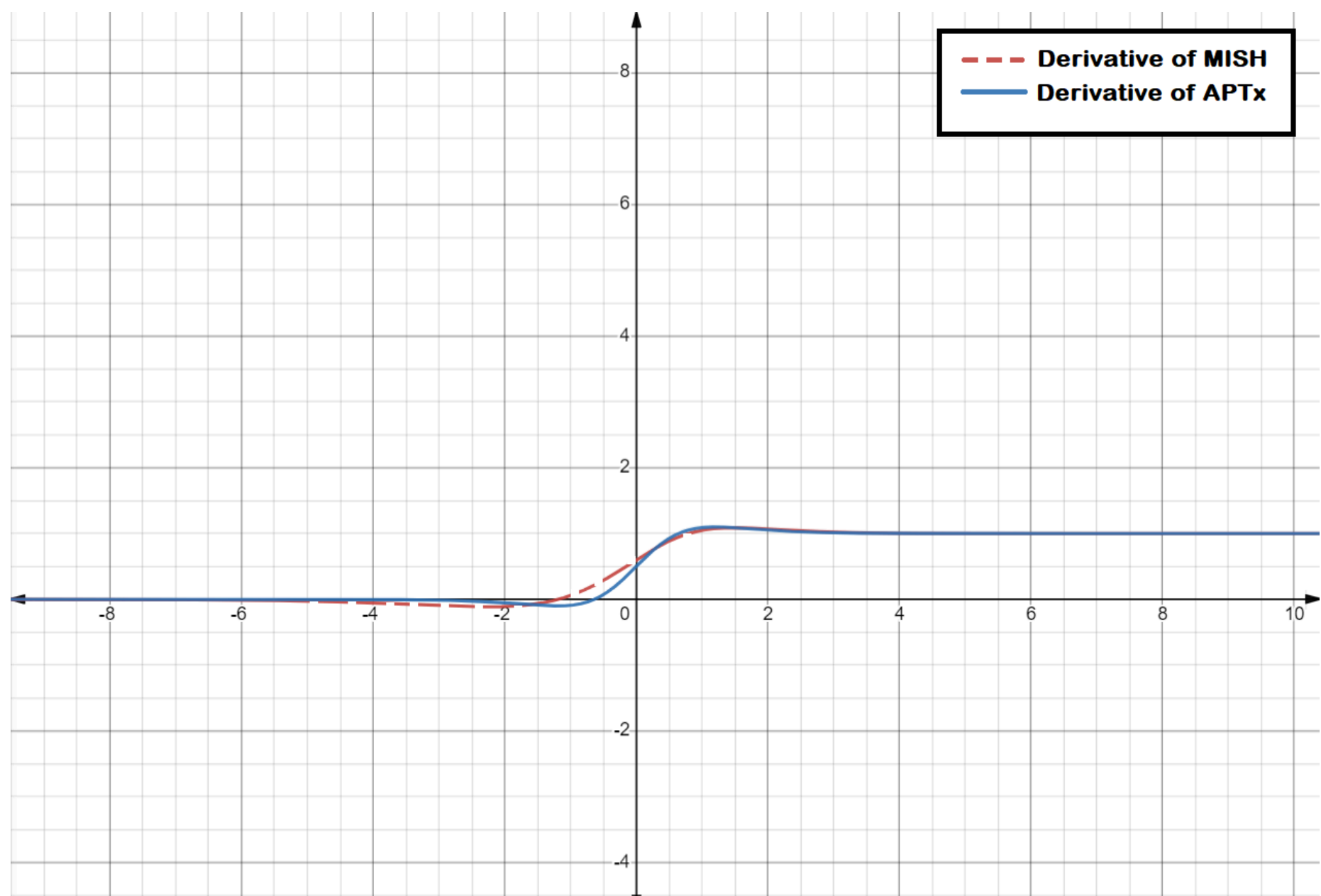

**Figure 6**: Graph of derivatives of MISH and APTx with $\alpha = 1$, $\beta = 1$ and $\gamma = ½$

Even more overlapping between MISH, and APTx derivatives can be generated by varying values for $\alpha$, $\beta$, and $\gamma$ parameters, such as $\alpha = 1$ , $\beta = ½$ and $\gamma = ½$ closely maps the negative domain part of the APTx with the negative part of MISH. When $\alpha = 1$ , $\beta = 1$ and $\gamma = ½$ the positive domain part of APTx closely maps with the positive part of MISH.

So, we can use $\alpha = 1$ , $\beta = ½$ and $\gamma = ½$ values for the negative part, and $\alpha = 1$ , $\beta = 1$ and $\gamma = ½$ for the positive part in case we want to closely approximate the MISH activation function.

Interestingly, APTx function with parameters $\alpha = 1$ , $\beta = ½$ and $\gamma = ½$ behaves like the SWISH(x, 1) activation function, and APTx with $\alpha = 1$ , $\beta = 1$ and $\gamma = ½$ behaves like SWISH(x, 2). That is, one can generate the SWISH(x, $\rho$) activation function using APTx activation function with parameters $\alpha = 1$ , $\beta = \rho/2$, and $\gamma = ½$.

The ReLU activation function can also be approximated using the APTx activation function by fixing $\alpha$ = 1, and γ = ½, with the approximation improving as $\beta$ increases and converging to ReLU in the limit $\beta \rightarrow \infty$. In practice, setting $\alpha$ = 1, $\beta \approx 10^6$, and γ = ½ already produces a close approximation of ReLU.

Our APTx activation function requires lesser computations in forward propagation and its derivative also needs lesser computations during backward propagation when compared with MISH activation function.

## 6 Conclusion

MISH has similar or even better performance than SWISH which is better than the rest of the activation functions. Our proposed activation function APTx behaves similar to MISH but requires lesser mathematical operations in calculating value in forward propagation, and derivatives in backward propagation. This allows APTx to train neural networks faster and be able to run inference on low-end computing hardwares such as neural networks deployed on low-end edge-devices with Internet of Things. Interestingly, using APTx one can generate the SWISH(x, ρ) activation function at parameters $\alpha$ = 1, $\beta$ = ρ/2, and γ = ½. Furthermore, choosing $\alpha$ = 1, $\beta \approx 10^6$, and γ = ½ yields a close approximation of the ReLU activation function.

## References

1. Vinod Nair, Geoffrey E. Hinton, Rectified Linear Units Improve Restricted Boltzmann Machines, 2010.
2. Agarap AF. Deep learning using rectified linear units (relu)”. arXiv preprint arXiv:1803.08375. 2018 Mar 22.
3. Glorot, Xavier, Bordes, Antoine, and Bengio, Yoshua. Deep sparse rectifier networks. In Proceedings of the 14th International Conference on Artificial Intelligence and Statistics. JMLR W&CP Volume, volume 15, pp. 315–323, 2011.
4. Sun, Yi, Wang, Xiaogang, and Tang, Xiaoou. Deeply learned face representations are sparse, selective, and robust. arXiv preprint arXiv:1412.1265, 2014.
5. Szandała, Tomasz. "Review and comparison of commonly used activation functions for deep neural networks." In *Bio-inspired neurocomputing*, pp. 203-224. Springer, Singapore, 2021.
6. Maas, Andrew L, Hannun, Awni Y, and Ng, Andrew Y. Rectifier nonlinearities improve neural network acoustic models. In ICML, volume 30, 2013.
7. Clevert DA, Unterthiner T, Hochreiter S. Fast and accurate deep network learning by exponential linear units (elus). arXiv preprint arXiv:1511.07289. 2015 Nov 23.
8. Ramachandran P, Zoph B, Le QV. Swish: a self-gated activation function. arXiv preprint arXiv:1710.05941. 2017 Oct 16;7(1):5.
9. Misra D. Mish: A self regularized non-monotonic activation function. arXiv preprint arXiv:1908.08681. 2019 Aug 23.